# Planning Under Continuous Time and Resource Uncertainty: A Challenge for AI

John Bresina, Richard Dearden,* Nicolas Meuleau,†
Sailesh Ramakrishnan,‡ David Smith and Rich Washington§
NASA Ames Research Center
Mail stop 269-2
Moffet Field, CA 94035-1000, USA

## Abstract

We outline a class of problems, typical of Mars rover operations, that are problematic for current methods of planning under uncertainty. The existing methods fail because they suffer from one or more of the following limitations: 1) they rely on very simple models of actions and time, 2) they assume that uncertainty is manifested in discrete action outcomes, 3) they are only practical for very small problems. For many real world problems, these assumptions fail to hold. In particular, when planning the activities for a Mars rover, none of the above assumptions is valid: 1) actions can be concurrent and have differing durations, 2) there is uncertainty concerning action durations and consumption of continuous resources like power, and 3) typical daily plans involve on the order of a hundred actions. This class of problems may be of particular interest to the UAI community because both classical and decision-theoretic planning techniques may be useful in solving it. We describe the rover problem, discuss previous work on planning under uncertainty, and present a detailed, but very small, example illustrating some of the difficulties of finding good plans.

## 1 THE PROBLEM

Consider a rover operating on the surface of Mars. On a given day, there are a number of different scientific observations or experiments that the rover could perform, and these are prioritized in some fashion (each observation or experiment is assigned a scientific value). Different observations and experiments take differing amounts of time and consume differing amounts of power and data storage. There are, in general, a number of constraints that govern the rovers activities:

- There are time, power, data storage, and positioning constraints for performing different activities. Time constraints often result from illumination requirement—that is, experiments may require that a target rock or sample be illuminated with a certain intensity, or from a certain angle.

- Experiments have setup conditions (preconditions) that must hold before they can be performed. For example, the rover will usually need to be at a particular location and orientation for each experiment and will need instruments turned on, initialized, and calibrated. In general, there may be multiple ways of achieving some of these setup conditions (e.g. different travel routes, different choice of cameras).

- The amount of power available varies according to the time of day, since solar flux is a function of the angle of the sun.

Given these constraints, the objective is to maximize scientific return for the rover—that is, find the plan with maximal utility. Unfortunately, for many rover activities, there is inherent uncertainty about the duration of tasks, the power required, the data storage necessary, the position and orientation of the rover, and environmental factors that influence operations, e.g., soil characteristics, dust on the solar panels, ambient temperature, etc.

For example, in driving from one location to another, the amount of time required depends on wheel slippage and sinkage, which varies depending on slope, terrain roughness, and soil characteristics. All of these factors also influence the amount of power that is consumed. The amount of energy collected by the solar panels during this traverse depends on the length of the traverse, but also on the an-

*Research Institute for Advanced Computer Science (RIACS)
†QSS Group Inc.
‡QSS Group Inc.
§RIACS

gle of the solar panels. This is dictated by the slope and roughness of the terrain.

Similarly, for certain types of instruments, temperature affects the signal to noise ratio and, hence, affects the amount of time required to collect useful data. Since the temperature varies depending on the time of day and the weather conditions, this duration is uncertain. The amount of power used depends upon the duration of the data collection. The amount of data storage required depends on the effectiveness of the data compression techniques, which ultimately depends on the nature of the data collected.

In short, this domain is rife with uncertainty. Plans that do not take this uncertainty into account usually fail miserably. In fact, it has been estimated that the 1997 Mars Pathfinder rover spent between 40% and 75% of its time doing nothing because of plan failure.

One way to attack this problem is to rely on real-time or reactive replanning when failures occur. While this capability is certainly desirable, there are several difficulties with exclusive reliance on this approach:

- Spacecraft and rovers have severely limited computational resources due to power limitations and radiation hardening requirements. As a result, it is not always feasible to do timely onboard replanning.

- Many actions are potentially risky and require pre-approval by mission operations personnel. Because of the cost and difficulty of communication, the rover receives infrequent command uplinks (typically one per day). As a result, each daily plan must be constructed and checked for safety well in advance.

- Some contingencies require anticipation; e.g., switching to a backup system may require that the backup system be warmed up in advance. For time critical operations such as orbit insertions or landing operations there is insufficient time to perform these setup operations once the contingency has occurred, no matter how fast the planning can be done.

For these reasons, it is sometimes necessary to plan in advance for potential contingencies—that is, anticipate unexpected outcomes and events and plan for them in advance.

The problem that we have just described is essentially a decision-theoretic planning problem. More precisely, the problem is to produce a (concurrent) plan with maximal expected utility, given the following domain information:

- A set of possible goals that may be achievable, each of which has a value or reward associated with it.

- A set of initial conditions, which may involve uncertainty about continuous quantities like temperature, energy available, solar flux, and position. This uncertainty is characterized by probability distributions over the possible values.

- A set of possible actions, each of which is characterized by:
  - a set of conditions that must be true before the action can be performed. (These may include metric temporal constraints as well as constraints on resource availability.)
  - an uncertain duration characterized by a probability distribution.
  - a set of certain and uncertain effects that describe the world following the action. Uncertain effects on continuous variables are characterized by probability distributions.

Decision-theoretic planning is already known to be quite hard both in theory [20] and in practice. However, there are some characteristics of this domain, which, when taken together, make this planning problem both difficult and different from the kinds of problems that have been studied in the past:

Time: actions take differing amounts of time and concurrency is often necessary.

Continuous outcomes: most of the uncertainty is associated with continuous quantities like time and power. In other words, actions do not have a small number of discrete outcomes.

Problem size: a typical daily plan for a rover will involve on the order of a hundred actions.

While we have described this scenario for a rover, this kind of problem is not limited to robotics or even space applications. For example, in a logistics problem, travel durations are influenced by both traffic and weather considerations. Fuel use is likewise influenced by these "environmental" factors. There are temporal constraints on the availability and delivery of cargo, as well as on the availability of both facilities and crew. There are also constraints on fuel loading and availability, and on maintenance operations.

## 2 PREVIOUS WORK

There has been considerable work in AI on planning under uncertainty. Table 1 classifies much of this work along the following two dimensions:

Representation of uncertainty: whether uncertainty is modeled strictly logically, using disjunctions, or is modeled numerically, with probabilities.

Observability assumptions: whether the uncertain outcomes of actions are not observable, partially observable, or fully observable.

Table 1: A classification of planners that deal with uncertainty. Planners in the top row are often referred to as conformant planners, while those in the other two rows are often referred to as contingency planners

|  | Disjunction | Probability |
|---|---|---|
| Non Observable | CGP [34]<br>CMBP [11, 1]<br>C-PLAN [10, 15]<br>Fragplan [18] | Buridan [19]<br>UDTPOP [26] |
| Partially Observable | SENSp [14]<br>Cassandra [28]<br>PUCCINI [16]<br>SGP [37]<br>QBF-Plan [30]<br>GPT [7]<br>MBP [2] | C-Buridan [12]<br>DTPOP [26]<br>C-MAXPLAN [21]<br>ZANDER [21]<br>Mahinur [25]<br>POMDP [8] |
| Fully Observable | WARPLAN-C [36]<br>CNLP [27] | JIC [13]<br>Plinth [17]<br>Weaver [5]<br>PGP [4]<br>MDP [8] |

We do not discuss this work in detail here. A survey of some of this work can be found in Blythe [6]. A more detailed survey of work on MDPs and POMDPs can be found in Boutilier, Dean and Hanks [8]. Instead we will focus on why this work is generally not applicable to the rover problem and what can be done about this.

There are a number of difficulties in attempting to apply existing work on planning under uncertainty to spacecraft or rovers. First of all, the work listed in Table 1 assumes a very simple model of action in which concurrent actions are not permitted, explicit time constraints are not allowed, and actions are considered to be instantaneous. As we said above, none of these assumptions hold for typical spacecraft or rover operations. These characteristics are not as much of an obstacle for Partial-Order Planning frameworks such as SENSp [14], PUCCINI [16], WARPLAN-C [36], CNLP [27], Buridan [19], UDTPOP [26], C-Buridan [12], DTPOP [26], Mahinur [25] and Weaver [5]. In theory, these systems could represent plans with concurrent actions and complex temporal constraints. The requirements for a rich model of time and action are more problematic for planning techniques that are based on the MDP or POMDP representations, satisfiability encodings, the graphplan representation, or state- space encodings. These techniques rely heavily on a discrete model of time and action. (See [33] for a more detailed discussion of this issue.) Although semi-Markov decision processes (SMDPs) [29] and temporal MDPs (TMDP) [9] can be used to represent actions with uncertain durations, they cannot model concurrent actions with complex temporal dependencies. The factorial MDP model has recently been developed to allow concurrent actions in an MDP framework. However, this model is limited to discrete time and state representations. Moreover, existing solution techniques are either too general to be efficient on real-world problems (e.g. Singh and Cohn [31]), or too domain-specific to be applicable to the rover problem (e.g. Meuleau et al. [22]).

A second, and equally serious, problem with existing contingency planning techniques is that they all assume that uncertain actions have a small number of discrete outcomes. For example, in the representation popularized by Buridan and C-Buridan, a rover movement action might be characterized as shown in Figure 1. In this representation, each arrow to a propositions on the right indicates a possible outcome of the action, along with the associated probability of that transition. To characterize where a rover could end up after a move operation, we have to list all the different possible discrete locations. We would need to do something similar to characterize power usage. For most spacecraft and rover activities this kind of discrete representation is impractical most of the uncertainty involves continuous quantities, such as the amount of time and power an activity requires. Action outcomes are distributions over these continuous quantities. There is some recent work using models with continuous states and/or action outcomes in both the MDP [3, 23, 24, 32] and POMDP [35] literature, but this has not yet been applied to SMDPs and has primarily been applied to reinforcement learning rather than planning problems.

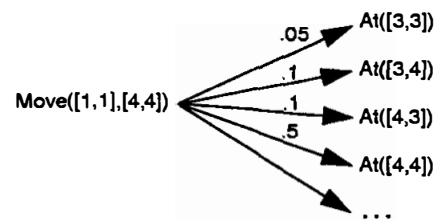

Figure 1: A C-Buridan action for movement.

Ultimately, the state that results from performing an action determines the future actions that will be taken, so some dimensions of an action's outcomes are discretized. However, this discretization is not a static property of the actions—instead, it depends on what goals or subgoals the planner is trying to accomplish. For example, suppose that the rover is trying to move to a certain location. If the objective is to place an instrument on a particular rock feature, then the tolerance in position is quite small. In contrast, if the objective is to take a picture from a different vantage point, then the tolerance can be significantly larger.

A third problem with conventional contingency planning technology is that it does not scale to larger problems. Part of the problem is that most of the algorithms attempt to account for all possible contingencies. In effect, they try to produce policies. For spacecraft and rover operations, this is not realistic or tractable—a daily plan can involve on the order of a hundred operations, many of which have uncertain outcomes that can impact downstream actions. The resulting plans must also be simple enough that they can be understood by mission operators, and it must be feasible to do detailed simulation and validation on them in a limited time period. This means that a planner can only afford to plan in advance for the "important" contingencies and must leave the rest to run-time replanning. Of the planning systems discussed above, only Just-In-Case (JIC) contingency scheduling [13] and Mahinur [25] exhibit a principled approach to choosing what contingencies to focus on. We will discuss this approach in more detail later.

## 3 A DETAILED EXAMPLE

In order to illustrate the problem further, in this section we give a detailed example of a very small rover problem. Figure 2 shows a "primary" plan and two potential branches. The primary plan consists of approaching a target point (VisualServo), digging the soil (Dig), backing up (Drive), and taking spectral images of the area (NIR). One potential alternate branch consists of replacing the spectral image with a high-resolution camera image of the target (Hi res). A second potential branch consists of taking a low-resolution panorama of the area (Lo res), performing on-board image analysis to find rocks in the panorama (Rock finder), and then taking spectral images of the rocks found (NIR). For this example, we assume that energy is only being depleted. (More generally, a rover would also be receiving energy input from charging.

Precedence constraints are denoted by arrows in the figure; for example, since HiRes can only be performed after Drive, there is an arrow from Drive to HiRes. For each action, there may be preconditions, expectations, and a local utility; in the figure, these appear above the plan actions. The preconditions specify under what conditions execution of the action may start. The expectations describe the expected resource consumption of the actions (in terms of mean and standard deviation); the relative width of distributions is illustrated graphically as well. The local utility is the reward received when the action terminates successfully: in this example, this will be when the preconditions are met and when the energy resource is non-negative at the end of execution.

In the example, consider the HiRes action. It has an energy precondition $E > 0.02$ Ah and a time precondition of 9:00 $\leq t \leq$ 16:00. The expected energy usage is 0.01 Amphours (Ah) with a standard deviation of 0 Ah (so in this case there is no uncertainty in the model). The expected duration is 5 seconds with a standard deviation of 1 second. The local utility of the action is $v = 10$.

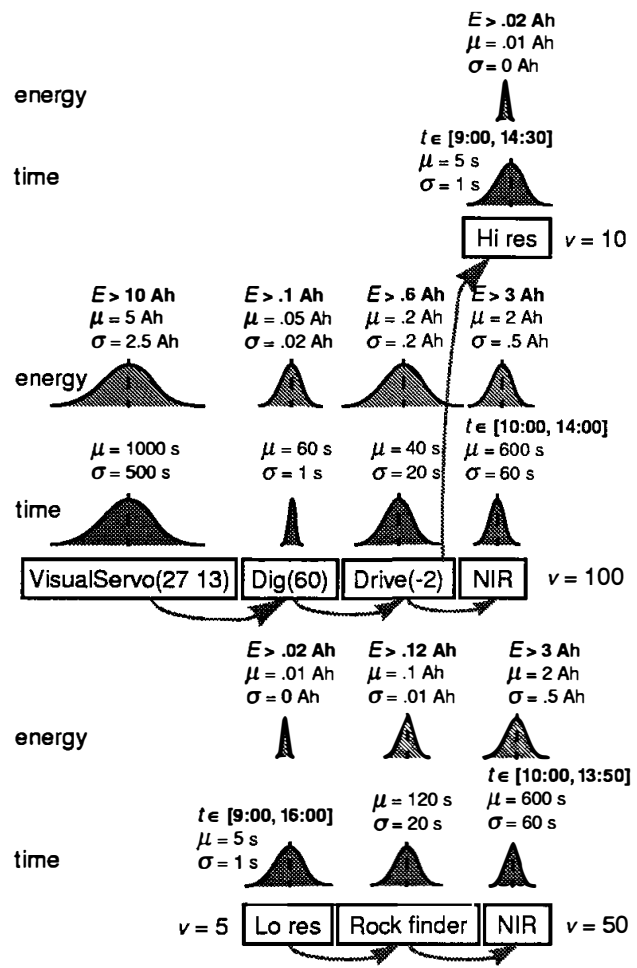

Figure 2: A detailed rover problem. A "main" plan, and two possible alternative branch plans are shown. Probability distributions for time and energy usage are shown for each action. Time and energy constraints for actions are shown in bold.

## 4 APPROACHES

There are several possible ways of attacking this problem of planning with continuous uncertain variables. In this section, we briefly discuss some of these, and the issues that arise.

### 4.1 COMPUTING THE OPTIMAL VALUE FUNCTION

Figure 3 shows the optimal value function for the problem in Figure 2. It represents the expected utility obtained by following an optimal (Markov) policy of the MDP repre-senting the problem. The figure was computed by working backwards from all possible activities that have positive reward and using dynamic programming to construct the optimal plan, after a fine discretization of both time and energy.[1] The curved hump where there is lots of power and time available corresponds to the primary plan, while the rectangular block corresponds to branching to the Rock finder plan and completing the NIR. The tail of the curved hump is a branch after the drive action to the HiRes plan. The flat surface with value 5 that covers nearly all the rest of the space is again an immediate branch to the Rock-Finder plan, but in this area there is not enough power or time to complete the plan, and only the LoRes reward is received. Figure 4 shows a cross-section through this surface for power equal to 11, showing how the various branches contribute to the overall plan. The utility curve of each branch, identified by its goal, represents the expected reward if we commit to the branch before knowing the initial conditions (start time). The maximum (upper envelope) of these curves is the expected utility of the best plan that first selects a branch depending on initial conditions, and then commits to this branch. The utility of the optimal policy (labeled as "all") is higher in some places than the utility of the best branch. This is because the optimal policy never commits prematurely to a branch. This increase in expected reward is due to the benefits of waiting to see how much time is available after part of the best plan has been performed, and branching to an alternative plan if the best one is unlikely to succeed in the remaining time, rather than comitting to a particular plan immediately.

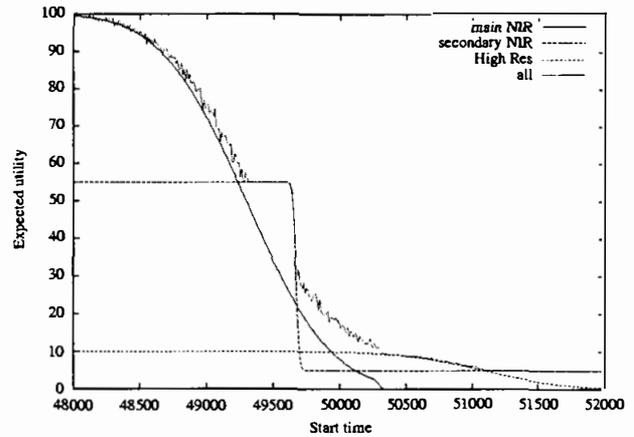

Figure 4: Slice of the optimal value function for energy = 11 Ah, along with the component curves that contribute to the overall utility.

Given a detailed contingent plan and the distributions for time and resource usage, it is relatively straightforward to evaluate the expected utility of the plan. If the distributions are very simple, it may be possible to compute this quantity exactly; more generally, this will have to be done with stochastic simulation. Thus, if we could generate all possible contingent plans for a problem, we could evaluate each of them and choose the one with highest utility. Of course this is completely impractical for problems of any size, partly because it is impossible to enumerate the conditions for conditional branches. The dynamic programming approach we took above is an alternative, but it too is computationally expensive, and it fails completely when resource availability is not monotonically decreasing (because optimization can no longer be performed through a single backward pass).

### 4.2 HEURISTIC APPROACHES

One possibility is to try to plan for the worst case scenario. Thus, in the example from the last section, we could assume that the drive operation requires time and power that is one or perhaps even two standard deviations above the mean. The trouble is, this approach is overly conservative and leads to plans with less science gain than is typically possible. In the example from the previous section, if plan execution was expected to begin at 13:45, this approach would lead us to build a "safe" primary plan that replaces NIR with the HiRes action, with expected utility of 10 in all cases, instead of the more ambitious current primary plan, with expected utility of 0 in the worst case, but 32 in the average case and 100 in the best case.

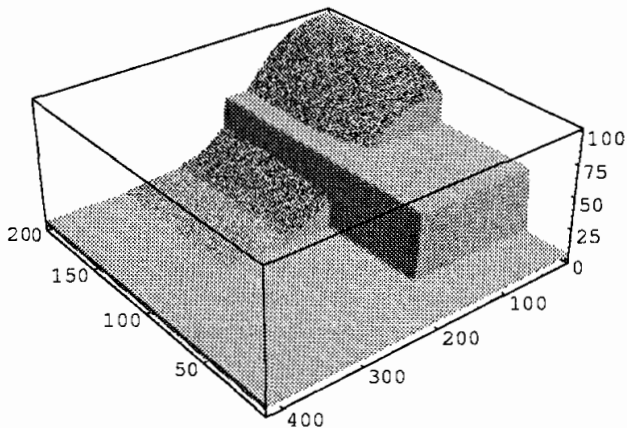

Figure 3: Optimal value function for the example in Figure 2. The left axis is increasing energy from 0 to 20. The right axis is start time from 14:30 down to 13:20. Vertical axis is expected utility.

A more ambitious approach to the problem would be to build an initial plan based on the expected behavior of various activities and then attempt to improve that plan by augmenting it with contingent branches. This is the approach taken by Drummond, Bresina and Swanson with their Just-in-Case (JIC) telescope scheduling [13]. This approach is intuitively simple and appealing, but extending it to problems like the one we have outlined is non-trivial. The primary difficulty is to decide where contingent branches should be added to a plan. In JIC scheduling, branches were added at the points with the greatest probability of plan failure. Given the distributions for time and resource usage this is relatively easy to calculate by statistical simulation of the plan. Unfortunately, the points most likely to fail are not necessarily the points where useful alternatives are available. The points of maximal failure probability may be too late in the plan to have enough time or power left for any useful alternative. A more efficient approach could be to identify the earliest point in time where we can predict with a given confidence that a failure is going to occur.

---

[1] With a grid of 420 steps for time and 200 steps for energy, the size of the state space is about $2.7 \cdot 10^6$. Moreover, it grows excponentially with the number of actions in the problem, so that this approach is unfeasible for any real size problem.

Unfortunately, the problem of finding "high utility" branch points is non-trivial. Figure 5 shows the expected utility over time of the possible plans with a single branch, for a fixed starting energy of 11. Note that at earlier start times, the plans with the highest expected utility are those that postpone the decision to later in the primary plan, where the possibility of receiving the 100 reward for the NIR action can be more accurately assessed. Between 49200 and 49700 seconds, the expected utility of 55 gained by immediately taking the RockFinder branch dominates as that plan is likely to succeed when started later than the primary plan. The value function for this branch drops off very sharply because there is relatively little uncertainty about the duration of this plan. As time advances, the value of branching later is apparent. Late branches look better when time is short because of the chance that an earlier action will happen unusually quickly, allowing the primary plan to be completed. Late branches to the RockFinder perform worse than to HiRes because there is rarely enough time remaining after the VisualServo action to complete the RockFinder plan. These plans finally dominate when there is very little time available because even the HiRes branch is unlikely to be completed.

### 4.3 FINDING THE BRANCH CONDITIONS

Once we have decided to add a branch to a plan, there is still a problem of deciding under what conditions to take the branch. Once again, we could use dynamic programming to compute the optimal conditions, but this suffers from the problems we described above. In addition, as Figure 3 illustrates, the optimal conditions can be extremely complex and hard to represent. The flat surfaces of utility 5 and 55 correspond to branching to the RockFinder plan before the first step of the primary plan. The primary plan (along with the later possible branch to the HiRes plan) is of higher expected utility where the surface is curved. The conditions for the branch point at the beginning of the primary plan are thus the boundaries between the curved surfaces and the flat surfaces. The boundaries are in this case discontinuous, corresponding to a disjunctive condition.

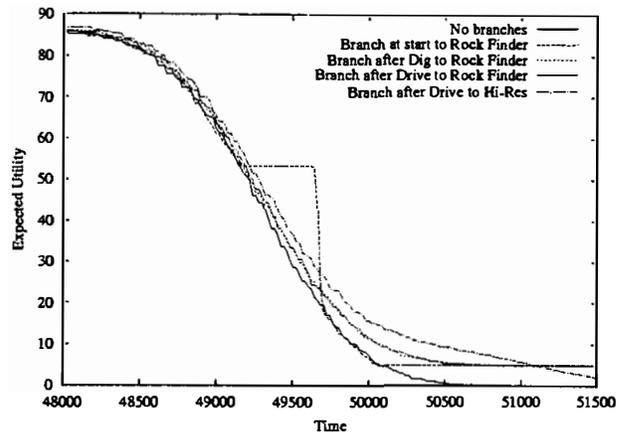

Figure 5: Utility for a single branch at different possible branch points with energy = 11.

It is important to bear in mind that the boundaries are generally places where the values of two different branches are equal, which means that approximate solutions will usually be acceptable here. One possibility is to treat the continuous dimensions of the problem as independent, which results in rectangular regions. This works well in most cases, but the boundaries must be chosen with care where there are abrupt edges in the value function. This approximation may also fail if there are dependencies between the dimensions, for example when the energy used for driving is dependent on the actual time spent, rather than being treated independently as in our example.

## 5 CONCLUSIONS

For a Mars rover, uncertainty is absolutely pervasive in the domain. There is uncertainty in the duration of many activities, in the amount of power that will be used, in the amount of data storage that will be required, and in the location and orientation of the rover. Unfortunately, current techniques for planning under uncertainty are limited to simple models of time, and actions with discrete outcomes. In the rover domain there is concurrent action, actions of differing duration, and most of the uncertainty is associated with continuous quantities like time, power, position and orientation.

For any non-trivial problem, it seems unlikely that exact or optimal solutions will be possible. Nor do we have good heuristic techniques for generating effective contingent plans. It seems that new and dramatically different approaches are needed to deal with this kind of problem.

**Acknowledgements**

Thanks to Tania Bedrax-Weiss, Jeremy Frank and Keith Golden for discussions on this subject and comments on drafts of the paper. This research was supported by NASA Ames Research Center and the NASA Intelligent Systems program.